\documentclass[10pt,twocolumn,letterpaper]{article}

\usepackage[accsupp]{axessibility}  

\usepackage[pagenumbers]{iccv} 

\usepackage{tcolorbox}
\usepackage{colortbl} 
\usepackage{xcolor}   
\definecolor{bgColor}{RGB}{250,227,212}

\definecolor{iccvblue}{rgb}{0.21,0.49,0.74}
\usepackage[pagebackref,breaklinks,colorlinks,allcolors=iccvblue]{hyperref}

\def\paperID{13140} 
\def\confName{ICCV}
\def\confYear{2025}

\title{MolParser: End-to-end Visual Recognition of Molecule Structures in the Wild}

\author{
    Xi Fang \quad Jiankun Wang \quad Xiaochen Cai \quad Shangqian Chen \quad
    Shuwen Yang \\ Haoyi Tao \quad Nan Wang \quad Lin Yao \quad
    Linfeng Zhang \quad Guolin Ke \\  
    DP Technology \\  
    {\tt\small \{fangxi, wangjiankun, caixiaochen, chenshangqian ,yangsw,} \\  
    {\tt\small taohaoyi, wangnan, yaol, zhanglf, kegl\}@dp.tech}  
}

\begin{document}
\maketitle
\begin{abstract}
In recent decades, chemistry publications and patents have increased rapidly. A significant portion of key information is embedded in molecular structure figures, complicating large-scale literature searches and limiting the application of large language models in fields such as biology, chemistry, and pharmaceuticals. The automatic extraction of precise chemical structures is of critical importance. However, the presence of numerous Markush structures in real-world documents, along with variations in molecular image quality, drawing styles, and noise, significantly limits the performance of existing optical chemical structure recognition (OCSR) methods. We present MolParser, a novel end-to-end OCSR method that efficiently and accurately recognizes chemical structures from real-world documents, including difficult Markush structure. We use an extended SMILES encoding rule to annotate our training dataset. Under this rule, we build MolParser-7M, a large-scale OCSR dataset based on our E-SMILES representation. While utilizing a large amount of synthetic data, we employed active learning methods to incorporate substantial in-the-wild data, specifically samples cropped from real patents and scientific literature, into the training process. We trained an end-to-end molecular image captioning model, MolParser, using a curriculum learning approach. MolParser significantly outperforms classical and learning-based methods across most scenarios, with potential for broader downstream applications. The dataset is publicly available in \href{https://huggingface.co/datasets/AI4Industry/MolParser-7M}{huggingface}.
\end{abstract}
    
\section{Introduction}
\label{sec:intro}

\begin{figure}[t]
  \centering
  \includegraphics[width=\linewidth]{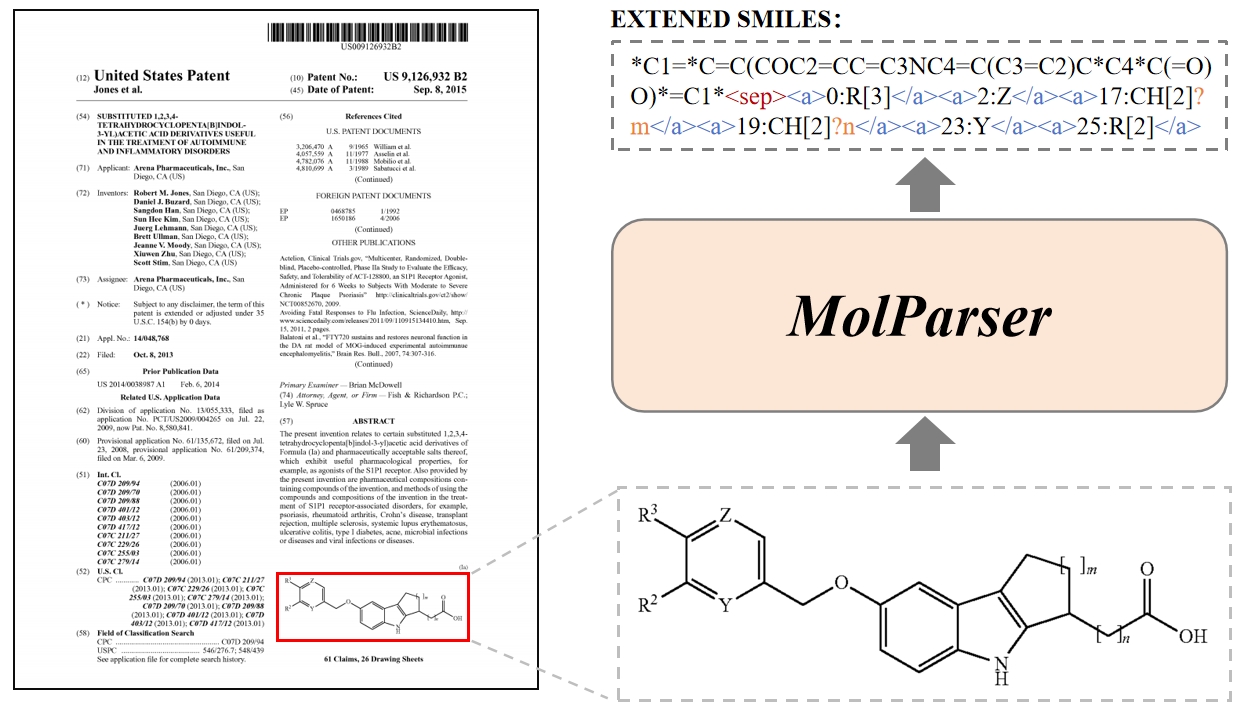}
   \caption{\textbf{MolParser} uses an end-to-end transformer to extract the chemical structures to string expression from the real patent or literature. We extend the SMILES format to enable the representation of more complex molecular structures including Markush.}
   \label{fig:overall}
\end{figure}

A significant portion of chemical information remains locked in unstructured formats within printed or digital documents, such as patents and scientific papers. In many of these documents, particularly in the field of chemistry, molecular structures are presented as images. These graphical depictions are essential for drug discovery, patent analysis, and chemical information retrieval. However, extracting them into machine-readable text remains a significant challenge. The automated interpretation of molecular structures from document images, a task known as Optical Chemical Structure Recognition (OCSR), is therefore of growing importance. With the rise of large language models (LLMs), increasing efforts are being directed toward applying them to the understanding of scientific literature. Converting molecular structure images into structured, interpretable text not only advances OCSR but also enables LLMs to more effectively process patents and scientific documents in chemistry-related domains.

OCSR aims to automatically convert chemical structure diagrams from scientific literature, patents, and other scanned documents into machine-readable string such as SMILES ~\cite{weininger1988smiles} representation. SMILES, though widely used for molecular representation, has notable limitations in handling complex chemical entities. It struggles with representing Markush structures, which are used in patents to describe a broad class of molecules by allowing variability at certain positions, enabling the protection of entire families of compounds. Additionally, SMILES cannot effectively handle connection points, abstract rings,  ring attachments with uncertainty position, duplicated structures or polymers, all of which require a level of flexibility that its linear format does not support. Furthermore, SMILES is not well-suited for tasks involving large models, such as Markush-molecule matching, as its structure lacks the clarity and hierarchical organization needed for efficient interpretation by machine learning models. These limitations hinder its utility in advanced cheminformatics applications.

\begin{figure}[t]
  \centering
  \includegraphics[width=1.0\linewidth]{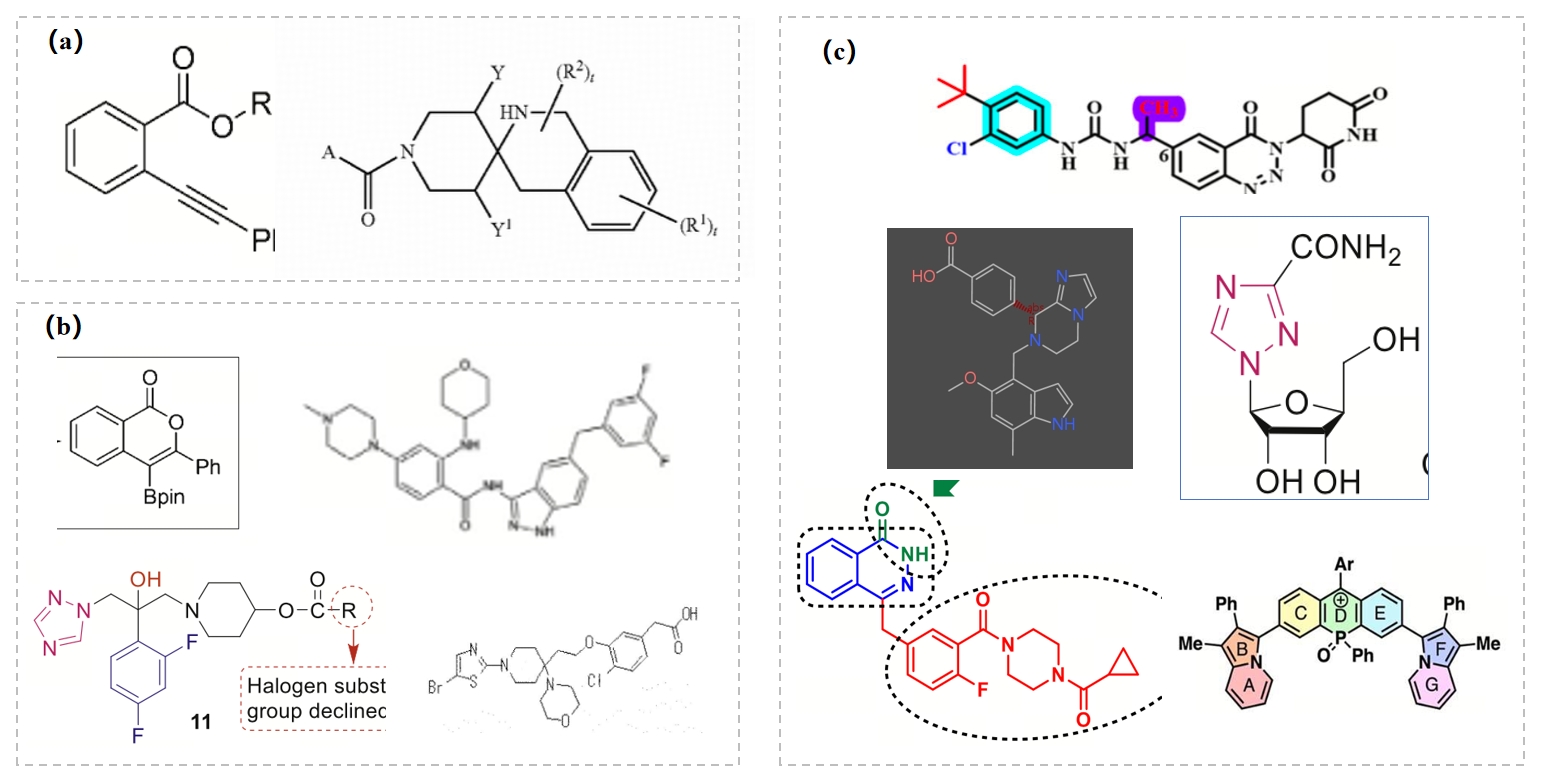}
   \caption{\textbf{In-the-wild problem in OCSR.} In real world patent and literature, we utilize object detection to locate and extract molecular images. However, there are several challenging cases that need to be addressed. These include (a) abbreviations and Markush structures, (b) image noise, blur and interference from surrounding elements, and (c) various drawing styles.}
   \label{fig:noise}
\end{figure}

The complexity of OCSR arises from the intricate nature of chemical diagrams, which not only include atoms and bonds but also various annotations, nested structures, and ring connections. These elements make traditional Optical Character Recognition (OCR) techniques inadequate for this task. Another challenge of the OCSR task lies in the varying styles and visual representations of chemical diagrams, which can differ in terms of drawing styles, colors, and formatting. These variations complicate the extraction of molecular structures. Additionally, during the recognition process, non-chemical elements such as surrounding text or other images from the paper may be mistakenly captured, further hindering the accurate identification of the chemical structures. These issues, combined with the noise, distortions, and font variations often found in document images, make the task even more complex.

Early OCSR approaches follow a graph reconstruction paradigm, where molecular structures were rebuilt by identifying key components such as atoms, bonds, and charges. These were typically extracted using hand-crafted image processing techniques, with rules applied to connect the elements and form a graph representation. While some recent methods have introduced deep learning for atom and bond detection, they still rely on hand-crafted rules ~\cite{filippov2009optical, smolov2011imago, oldenhof2020chemgrapher, xu2022molminer, zhang2022abc} for linking the recognized components. There are also some new methods that utilize Graph Neural Networks (GNNs) or transformers as replacements for traditional rule-based approaches~\cite{morin2023molgrapher, chen2024molnextr}. Constructing molecular structures uses graph representations and ultimately derives the string representation of these molecules (e.g. SMILES). This multi-step process makes training and model fine-tuning relatively complex and limits the robustness of these methods when dealing with noise and distortions commonly found in real-world documents and patents. As a result, their performance in handling noisy data remains suboptimal.

With the recent advancements in deep learning, end-to-end neural network models have become the dominant approach for OCSR. These methods combine image recognition with sequence generation tasks, directly converting input images into string molecular representations ~\cite{clevert2021img2mol, xu2022swinocsr, rajan2023decimer}. This method can be seen as a special case of image caption. However, despite significant progress, current models face real-world challenges, particularly when processing complex, noisy, or previously unseen chemical structures in patent documents. An important reason for this is that the training data utilized by these methods is often small-scale synthetic data, which differs significantly from real-world data scenarios. This highlights the need for developing more robust and versatile OCSR systems capable of handling greater diversity and complexity in real-world applications.

In order to better address the challenges of the OCSR task in real-world literature. We introduce a new end-to-end framework named MolParser for Optical Chemical Structure Recognition in the wild, illustrated in Figure \ref{fig:overall} The main contributions of this paper are:

\textbf{Extended SMILES}
We extend the SMILES representation to accommodate a broader range of specialized molecules commonly found in patents and literature, including Markush structures, connection points, abstract rings, ring attachments with uncertainty position, duplicated structures, and polymer structures. Additionally, this extended SMILES format is compatible with RDKit ~\cite{rdkit} and is also LLM-friendly, making it convenient for using LLMs to perform various analyses and processing on molecules.

\textbf{MolParser-7M Datasets}
Based on our extended SMILES representation, we construct MolParser-7M , the largest annotated molecular recognition training dataset to our knowledge, with over 7 million paired image-SMILES data. MolParser-7M contains a large amount of diverse synthetic data, as well as in-the-wild data (cropped from real-world PDF scans). Additionally, we design a human-in-the-loop data engine to extract the most training-relevant molecular images from millions of patents and scientific papers, followed by meticulous manual annotation and cross review. We also provide a new OCSR benchmark, WildMol, including 10k ordinary molecules (WildMol-10k) and 10k Markush structures (WildMol-10k-M). All the samples are annotated by our extended SMILES (E-SMILES) format.

\textbf{MolParser Model}
We regard OCSR tasks as a special type of image captioning task, where the content of the caption is an extended SMILES string. We develop MolParser model using an end-to-end image caption architecture, which includes a vision encoder, a feature compressor, and a BART \cite{lewis2019bart} decoder to generate extended SMILES strings. We employed curriculum learning to train the MolParser model, first pretraining it on the diverse synthetic data of the MolParser7M dataset, gradually increasing the intensity of data augmentation during training. Afterward, we fine-tuned the model on a subset containing 400k in-the-wild real data. As a result, on the WildMol-10k benchmark, MolParser achieved a state-of-the-art accuracy of 76.9\%, significantly outperforming existing methods, with MolScribe \cite{qian2023molscribe} at 66.4\% and MolGrapher \cite{morin2023molgrapher} at 45.5\%. Additionally, with an inference speed of up to 40 FPS (131 FPS for the tiny version), MolParser is better suited for industrial applications compared to existing methods.

\section{Related Works}
\label{sec:related}
Related work includes various molecular representation methods, such as SMILES, as well as algorithms for Optical Chemical Structure Recognition (OCSR).

\textbf{SMILES variants.}
The Simplified Molecular Input Line Entry System (SMILES) provides a highly compact, linear string representation of molecular structures by encoding atoms and bonds efficiently. Its conciseness and simplicity have made SMILES a widely adopted standard in cheminformatics for molecular storage, retrieval, and similarity assessments. SMILES notation represents molecules, but cannot depict molecular templates like Markush structures. FG-SMILES suggested in Image2SMILES \cite{khokhlov2022image2smiles} attempts to solve this problem. This is an extension of standard SMILES, where a substituent or R-group can be written as a single pseudo-atom. However, this approach has limited scalability, as it struggles to support abstract rings, ring attachments with uncertainty position, duplicated structures, and polymer structures. At the same time, it is difficult to ensure compatibility with the current leading molecular processing tool, RDKit ~\cite{rdkit}, which complicates subsequent processing and analysis.

\textbf{Image captioning based OCSR (End-to-End).}
Most recent end-to-end deep learning approaches leverage image captioning techniques, which involve generating descriptive textual representations of images. These models employ an encoder to extract visual features from images and a decoder to convert them into SMILES \cite{weininger1988smiles} or InChI \cite{heller2013inchi} sequences. Specifically, models like MSE-DUDL \cite{staker2019molecular}, DECIMER \cite{rajan2020decimer}, Img2Mol \cite{clevert2021img2mol}, ChemPix \cite{weir2021chempix}, and MICER \cite{yi2022micer} utilize a convolutional encoder paired with various recurrent decoders (RNN, GRU or LSTM). Subsequent works have introduced transformer-based encoder-decoder architectures, such as DECIMER 1.0 \cite{rajan2021decimer}, DECIMER 2.0 \cite{rajan2023decimer}, SwinOCSR \cite{xu2022swinocsr}, IMG2SMI \cite{campos2021img2smi}, Image2SMILES \cite{khokhlov2022image2smiles} and Image2InChI\cite{li2024image2inchi}. The advantage of these algorithms lies in their fast end-to-end speed and strong generalization performance. But a significant drawback of these image captioning methods is their requirement for large training datasets. Most of these methods rely on generated data and do not achieve satisfactory performance in in-the-wild scenarios such as patents or literature.

\textbf{Graph-based OCSR (Atom-Bond).}
Traditional OCSR methods rely on hand-crafted image processing rule to detect molecular components and reconstruct the molecular graph \cite{mcdaniel1992kekule, bukhari2019chemical, filippov2009optical, ouyang2011chemink, sadawi2012chemical, smolov2011imago, Peryea2022Molvec}. Recent approaches utilize deep learning for component detection or segmentation \cite{oldenhof2020chemgrapher, xu2022molminer, zhang2022abc}. While more recent utilize deep learning to build the graphs instead of using hand-crafted rule \cite{qian2022robust, yoo2022image}. 
However, this approach is complex and computationally slow, and its complexity makes extensive manual labeling nearly impossible, resulting in a heavy reliance on generated data. This reliance, in turn, makes it vulnerable to noise interference in real-world applications and contributes to lower generalization performance. Even though the latest MolGrapher \cite{morin2023molgrapher} achieves state-of-the-art results on several benchmarks, it still encounters challenges in real-world literature scanning scenarios. Similar methods include MMSSC-Net \cite{zhang2024mmssc} and MolScribe \cite{qian2023molscribe}.

\textbf{Markush Recognition.}
Markush structures are chemical representations that use variable groups to describe a family of related compounds. They are commonly found in patents to claim broad molecular classes. These structures often contain R-groups, repeating units, and variable attachments. Markush recognition \cite{haupt2009markush, beard2020chemschematicresolver, wang2022multi, morin2025markushgrapher} remains a major challenge. Moreover, existing tools such as RDKit \cite{rdkit} cannot parse Markush structures. They do not support inputs with undefined groups or variable atoms. These tools also fail to render valid images of Markush structures. As a result, generating large-scale training data becomes more difficult. The lack of standardized representation and visualization limits data augmentation. This creates a bottleneck for training robust models on Markush recognition.

\section{MolParser}
\label{sec:method}

\begin{figure}[t]
  \centering
  \includegraphics[width=1.0\linewidth]{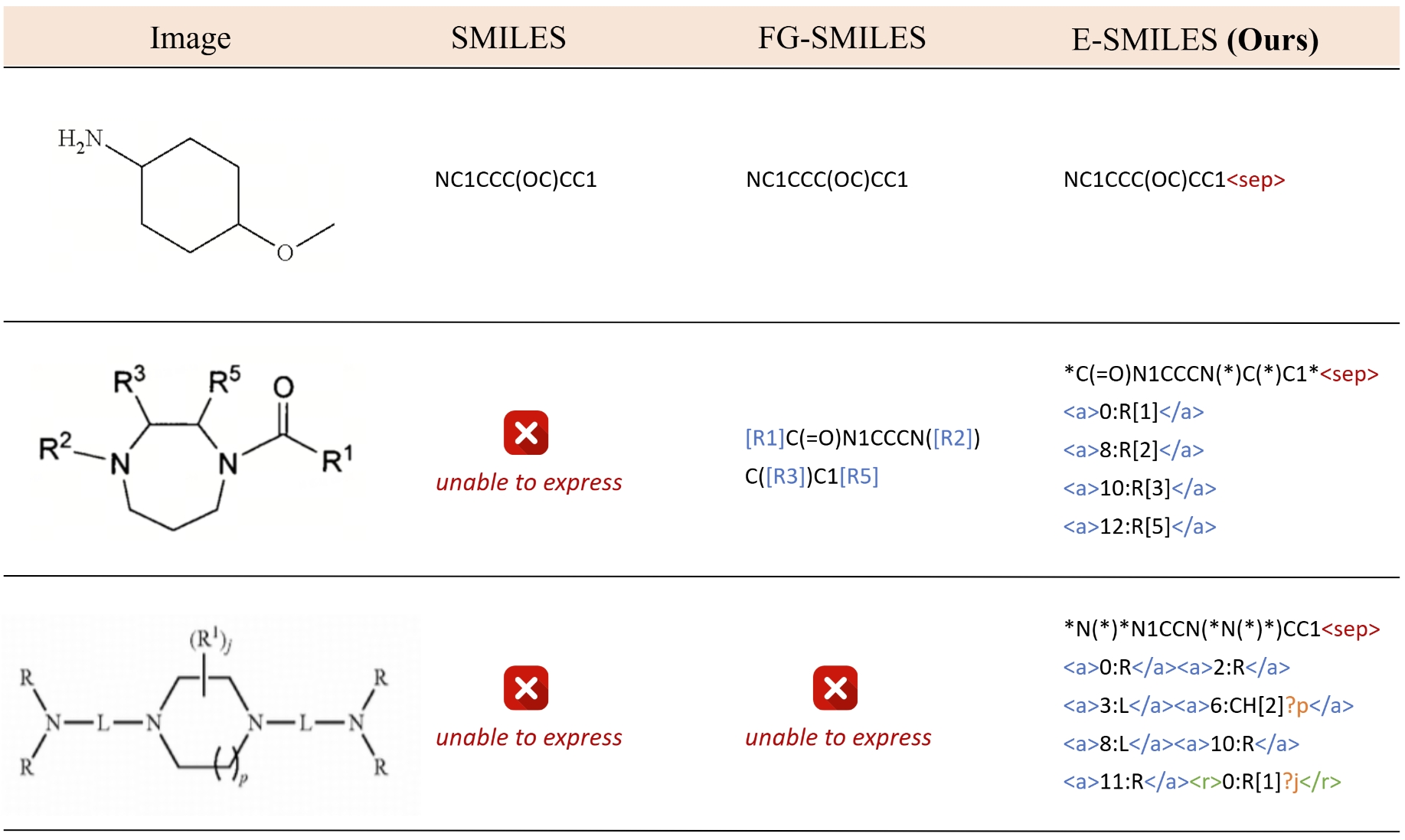}
   \caption{\textbf{Comparison of molecular expressions.} Our extended SMILES is able to express more complex Markush structures.}
   \label{fig:smiles}
\end{figure}

\subsection{Extended SMILES}
\label{subsec:extsmiles}
We extend the SMILES representation method, abbreviated as E-SMILES, to more effectively represent the Markush structures commonly found in patents, as well as complex compositions such as connection points, abstract rings, ring attachments with uncertainty position, duplicated structures, and polymer structures. Additionally, we ensure that this approach is compatible with RDKit and LLM-friendly, facilitating subsequent analysis and processing tasks. The extended SMILES format will be denoted as:
\texttt{SMILES<sep>EXTENSION}. Where \texttt{SMILES} refers to the RDKit-compatible SMILES representation. Special token \texttt{<sep>} serves as the separator, and optional \texttt{EXTENSION} represents supplementary descriptive used to handle complex cases such as Markush structures.

In \texttt{EXTENSION}. We use some XML-like special tokens to represent certain special functional groups. For Markush R-groups and abbreviation groups, we add special tokens \texttt{<a>} and \texttt{</a>} to encapsulate descriptions of these special substituents. Similarly, we use special tokens \texttt{<r>} and \texttt{</r>} for ring attachments with uncertainty position; \texttt{<c>} and \texttt{</c>} for abstract rings. Additionally, there is a special token \texttt{<dum>} representing a connection point. For the specific description of each functional group, we use the following format: \texttt{[INDEX]:[GROUP\_NAME]}. Figure \ref{fig:smiles} shows a example of our extended SMILES. Although there are numerous complex Markush structures in actual patents, our extended SMILES 
 (E-SMILES) rules can still adequately address these cases. For more details, please refer to the supplementary materials \ref{sup:extended_smiles}.

\subsection{Architecture}
\label{subsec:model}
Motivated by image captioning approach ~\cite{li2023trocr, xu2022swinocsr, lee2023pix2struct, blecher2023nougat}, we employ a single transformer only architecture to translate molecular structure images into our extended SMILES format. Our MolParser model has three components, an image encoder, a feature compresser, and a SMILES decoder.

We use an ImageNet \cite{deng2009imagenet} pretrained Swin-Transformer \cite{liu2021swin} as the image encoder for our MolParser. Similar to LLaVA \cite{liu2024llava}, we employ a two-layer MLP as a vision-language connector to compress the feature dimension. Finally, we adopt a decoder-only Transformer architecture, BART-Decoder \cite{lewis2019bart}, to decode the compressed image features and autoregressively generate the extended SMILES (E-SMILES) sequence via next-token prediction.

\subsection{Training}
\label{subsec:train}
We employ autoregressive algorithm to generate the E-SMILES sequence. For paired training data, we first use a tokenizer to convert the E-SMILES sequence into a token sequence. Each token generally represents an atom, an abbreviated group name, a number, a special symbol, or a special token defined in E-SMILES. The training process is conducted in two stages: pretraining stage and supervised finetuning stage.

In the first stage, we conduct pretraining using a synthetic dataset comprising over 7.7 million paired training samples. Concurrently, we employ a curriculum learning \cite{bengio2009curriculum} approach during pretraining to achieve better convergence. In the early phase of pretraining, we refrain from utilizing data augmentation and restrict our focus to simple molecules with a SMILES token count of fewer than 60. Subsequently, we progressively increase the intensity of data augmentation and incorporate molecules with longer token sequences into the training process. The detail of training dataset is described in Section \ref{sec:datasets}.

In the second stage, we fine-tune the model with a set of human-annotated in-the-wild data. The purpose of this stage is to further enhance the generalization ability of MolParser in the real scene. Previous methods \cite{clevert2021img2mol, morin2023molgrapher, rajan2023decimer} only use synthetic data, which exhibited a relatively restricted distribution in images and consisted of relatively simple molecular structures, resulting in suboptimal performance in real-world applications. In contrast, we employ an active learning approach to construct a data engine that extracted approximately 400,000 molecular images deemed most valuable for model learning from over 1.22 million real patents and academic papers, supplemented by manual annotation and secondary review. After incorporating these data for fine-tuning, we further improve MolParser’s ability in real application. The detail of our active leanring data engine is described in Section \ref{sec:engine}.

\section{MolParser Data Engine}
\label{sec:engine}

As paired OCSR training data is not abundant on the Internet, we build a data engine to enable the collection of our 7.7 million OCSR paired dataset, MolParser-7M, with low cost. The data engine has two stages: (1) start up phase with fully synthetic training data, (2) active learning with human in the loop. Here is the detail:

\textbf{Start up with synthetic data.} Due to the significant difficulty in annotating molecular data in SMILES format, the cost of manual annotation in large volumes is quite high. Our analysis shows that, on average, it takes over three minutes for an expert to annotate a single molecule from scratch. Therefore, we chose to initiate the data engine with synthetic data and gradually expand it to real data. We utilized a diverse range of molecular structure sources and various image rendering methods, generating over 7 million pairs of images and extended-SMILES. Details of the synthetic data can be referenced in Section \ref{sec:datasets}.  Then we can use synthetic data to train the initial version of our MolParser model.

\textbf{Active Learning with Human in the Loop.} To further enhance the generalization performance of our model, we extract molecular training data directly from patents and scientific literature in PDF format and manually annotate the molecular structures. Initially, we train a YOLO11 ~\cite{yolo11} object detection model, named as MolDet, to locate molecules within these documents. We use 1.22 million real PDF files, including patent documents from various international patent offices and open-access papers from Internet such as bioRxiv, medRxiv, and ChemRxiv. From this dataset, we extract over 20 million molecular images. After employing image p-hash similarity analysis to remove duplicate or highly similar molecular images, this number reduces to 4 million. Due to the sheer volume of this dataset, manual annotation of all images is not feasible. Therefore, we introduce active learning algorithms to identify and select samples with higher importance.

We perform 5-fold training on synthetic data to obtain five distinct models. Each model independently generates predictions, resulting in five E-SMILES strings per molecule image. We extract the standard SMILES sequences before the \texttt{<sep>} token and compute the pairwise Tanimoto similarity, taking the average similarity score as the confidence score for OCSR prediction. We observe that data with low confidence scores often correspond to images with poor quality, whereas data with high confidence scores typically indicate a high probability of correct predictions. We then randomly select samples with confidence scores between 0.6 and 0.9, as we believe these samples present significant recognition challenges and are highly beneficial for training. We manually annotate the molecular structures for these selected samples.

During the annotation process, we use model predictions as pre-annotations, which annotators modify as needed, without editing molecular structures from scratch. Two different annotators independently review the model predictions or manual modifications to verify their correctness. Analysis shows that leveraging pre-annotations reduces annotation time per molecule to 30 seconds, achieving approximately 90\% savings in manual labor compared to annotation from scratch.

After every 80,000 completed annotations, we incorporate the data into the training set, update the 5-fold models, and repeat the active learning cycle. This iterative process enhances both the model's generalization ability and the quality of the pre-annotated data. Through this loop, we construct a dataset of 400,000 manually annotated images.

In the final dataset, 56.04\% of annotations directly use model pre-annotations, 20.97\% pass review after a single manual correction, 13.87\% are accepted after a second round of annotation, and 9.13\% require three or more rounds of annotation.

\section{MolParser-7M Dataset}
\label{sec:datasets}

Our dataset, MolParser-7M, consists of 7.7M diverse molecule structure images. As far as we know, MolParser contains the largest number of paired samples in open-sourced OCSR datasets and it is the only open-source training set that includes a significant amount of real molecule images croped from real patent and literature. The largest open-access OCSR paired dataset available before this work was MolGrapher-300K \cite{morin2023molgrapher}, which included only 300k paired samples, all of which are synthetic data.

\textbf{Synthetic Training Data Generation.}
To launch our data engine, we first generated approximately 7M of paired OCSR training data. To obtain a more diverse distribution of data, we collected a substantial number of molecular structures from various sources and additionally generated a significant number of Markush structures at random. The sources from which we obtained molecular structure data include: ChEMBL \cite{gaulton2012chembl} database, PubChem \cite{kim2019pubchem} database, Kaggle BMS \cite{bms} dataset. Training images are then generated from SMILES using the molecule drawing library RDKit \cite{rdkit} and epam.indigo \cite{indigo}. Similar to previous work MolGrapher \cite{morin2023molgrapher}, in order to increase image diversity, rendering parameters are also randomly set. The specific data sources are listed in the Table~\ref{tab:pretrain}.

\begin{table}[t]\scriptsize
\renewcommand{\arraystretch}{1.2}

\begin{subtable}{0.47\textwidth}
    \centering
    \setlength{\tabcolsep}{0.9mm} 
    \begin{tabular}{p{2.1cm}|c|p{5.2cm}}

         \rowcolor{bgColor}
         subset & ratio & source \\
         \hline
         Markush-3M & 40\% & random groups replacement from PubChem \cite{kim2019pubchem} \\
         \hline
         ChEMBL-2M & 27\% & molecules selected from ChEMBL \cite{gaulton2012chembl} \\
         \hline
         Polymer-1M & 14\% & random generated polymer molecules \\
         \hline
         PAH-600k & 8\% & random generated fused-ring molecules \\
         \hline
         BMS-360k & 5\% & molecules with long carbon chains from BMS \cite{bms} \\
         \hline
         MolGrapher-300K & 4\% & training data from paper MolGrapher \cite{morin2023molgrapher} \\
         \hline
         Pauling-100k & 2\% & Pauling-style images drawn using epam.indigo \cite{indigo} \\
         \hline
    \end{tabular}
    \centering
        \caption{Datasets used in pretraining stage.
    }
\label{tab:pretrain}
\end{subtable}

\begin{subtable}{0.47\textwidth}
    \centering
    \setlength{\tabcolsep}{0.9mm} 
    \begin{tabular}{p{2.1cm}|c|p{5.2cm}}

         \rowcolor{bgColor}
         subset & ratio & source \\
         \hline
         MolParser-SFT-400k & 66\% & manually annotated data obtained using data engine \\
         \hline
         MolParser-Gen-200k & 32\% & synthetic data selected from pretraining stage \\
         \hline
         Handwrite-5k & 1\% & handwritten modelcules selected from Img2Mol \cite{clevert2021img2mol} \\
         \hline
    \end{tabular}
    \centering
        \caption{Datasets used in fine-tuning stage.
    }
\label{tab:finetuning}
\end{subtable}

\caption{\textbf{Summary of datasets used in MolParser training.} 
To construct MolParser-7M dataset, we use a very wide range of data sources.
}
\label{tab:dataset}
\end{table}

\textbf{Fine-tuning dataset construction.} We obtained approximately 400,000 manually annotated training data from the active learning data engine. In addition, we found that it is necessary to keep a part of synthetic data in the fine-tuning stage. To support handwritten molecule recognition, we also add some manually annotated handwritten molecules. The specific composition of fine-tuning data can be referred to Table~\ref{tab:finetuning}.

To assess the performance of the OCSR model in in-the-wild scenarios, we have also released an open-source OCSR test set, WildMol, comprising 20,000 human annotated molecule samples cropped from real PDF files. It presents greater difficulty and features an in-the-wild distribution compared to other open-source evaluation benchmarks.

\section{Experiments}
\label{sec:exp}

\begin{table*}[!ht]
\centering%
\resizebox{0.92\linewidth}{!}{
\begin{tabular}{lccccccc}
\toprule
\textbf{Method} & \begin{tabular}[c]{@{}c@{}}USPTO\\ (5719)\end{tabular} & \begin{tabular}[c]{@{}c@{}}UoB\\ (5740)\end{tabular} & \begin{tabular}[c]{@{}c@{}}CLEF\\  (992)\end{tabular} & \begin{tabular}[c]{@{}c@{}}JPO\\  (450)\end{tabular} & \begin{tabular}[c]{@{}c@{}}ColoredBG\cite{xiong2024alphaextractor}\\  (200)\end{tabular}  & \begin{tabular}[c]{@{}c@{}}USPTO-10K\cite{morin2023molgrapher}\\ (10000)\end{tabular} & \begin{tabular}[c]{@{}c@{}}\textbf{WildMol-10K}\\ (10000)\end{tabular} \\ \hline
\textit{Rule-based methods} \\
OSRA 2.1 \cite{filippov2009optical} \textsuperscript{$\ast$} & 89.3 & 86.3 & \textbf{93.4} & 56.3 & 5.5 & 89.7 & 26.3 \\
MolVec 0.9.7 \cite{Peryea2022Molvec} \textsuperscript{$\ast$} & 91.6 & 79.7 & 81.2 & 66.8 & 8.0 &  92.4 & 26.4 \\
Imago 2.0 \cite{smolov2011imago} \textsuperscript{$\ast$} & 89.4 & 63.9 & 68.2 & 41.0 & 2.0  & 89.9 & 6.9 \\ \hline
\textit{Only synthetic training} \\
Img2Mol \cite{clevert2021img2mol} \textsuperscript{$\ast$} & 30.0 & 68.1 & 17.9 & 16.1 & 3.5 & 33.7 & 24.4 \\
MolGrapher \cite{morin2023molgrapher} \textsuperscript{\textdagger $\ast$} & 91.5 & \textbf{94.9} & 90.5 & 67.5 & 7.5 & 93.3 & 45.5 \\ \hline
\textit{Real data finetuning} \\
DECIMER 2.7 \cite{rajan2023decimer} \textsuperscript{$\ast$} & 59.9 & 88.3 & 72.0 & 64.0 & 14.5 & 82.4 & 56.0 \\
MolScribe \cite{qian2023molscribe} \textsuperscript{$\ast$} & \textbf{93.1} & 87.4 & 88.9 & \underline{76.2}  & 21.0 & \textbf{96.0}  & 66.4 \\
\textbf{MolParser-Tiny}  (Ours)  & \underline{93.0} & 91.6 & \underline{91.0} & 75.6 & \textbf{58.5} & 89.5 & 73.1 \\
\textbf{MolParser-Small}  (Ours)  & \textbf{93.1} & 91.1 & 90.8 & \underline{76.2} & \underline{57.0} & \underline{94.8} & \underline{76.3} \\
\textbf{MolParser-Base}  (Ours) & \underline{93.0} & \underline{91.8} & 90.7 & \textbf{78.9} & \underline{57.0} & 94.5 & \textbf{76.9} \\
\bottomrule
\end{tabular}}\vspace{-1mm}
\caption{\textbf{Comparison of our method with existing OCSR models.} We report the accuracy. We use \textbf{bold} to indicate the best performance and \underline{underline} to denote the second-best performance. $\ast$: re-implemented results. \textdagger: results from original publications.}
\label{tab:results}
\end{table*}

In this section, we conduct extensive experiments on various OCSR benchmarks. Moreover, we demonstrate the application of our MolParser method in downstream tasks and reveal an intriguing finding: the image encoder of our MolParser model shows promising utility in the field of molecular property prediction.

\subsection{Evaluation datasets and metrics}
\label{exp_metric}

To compare our method with previous state-of-the-art approaches, we evaluate the model on several classic publicly available benchmarks, including USPTO \cite{filippov2009optical}, Maybridge UoB \cite{sadawi2012chemical}, CLEF-2012 \cite{piroi2010clef}, and JPO \cite{fujiyoshi2011robust}. However, these classic OCSR evaluation datasets are limited in size and contain systematic biases and annotation noise. To further assess the performance of our model, we conduct tests on a small but challenging in-the-wild OCSR dataset, ColoredBG \cite{xiong2024alphaextractor}, as well as a larger-scale dataset, USPTO-10k \cite{morin2023molgrapher}, containing 10,000 molecular images. Additionally, we evaluate model performance on our proposed WildMol dataset to comprehensively test the OCSR algorithm's performance in in-the-wild literature scenarios. WildMol-10K contains 10,000 regular molecules, and WildMol-10K-M contains 10,000 Markush structures. For evaluation metric, we use the classic accuracy metric, which is commonly applied in such tasks.

\subsection{State-of-the-art comparison}
\label{bench}

Table \ref{tab:results} compares OCSR methods across different benchmarks, where our MolParser method consistently outperforms existing approaches, including the previous state-of-the-art, MolGrapher \cite{morin2023molgrapher} and MolScribe \cite{qian2023molscribe}. On classical benchmarks such as USPTO, Maybridge UoB, JPO, ColoredBG and USPTO-10K, MolParser achieves satisfactory results. On our newly proposed, significantly more challenging test set, WildMol-10K, which consists of molecule images cropped from real patent literature, MolParser also demonstrates substantial improvements, confirming its ability to handle diverse molecular image data from various document sources.

We built a series of MolParser models with various sizes by using visual backbones of different scales, input resolutions, and BART decoders with varying parameter counts. As shown in Table \ref{tab:size}, our model demonstrates clear advantages in both speed and accuracy compared to previous state-of-the-art model MolGrapher, achieving a significantly better Pareto frontier. The throughout is tested in RTX-4090D. The Tiny variant of MolParser achieves a parsing speed of over 130 molecular images per second with minimal accuracy loss, enabling rapid extraction and parsing of molecular structures in ultra-large-scale unstructured documents. The Base variant of MolParser achieves the highest accuracy with a recognition speed of 40 molecular images per second. Due to its end-to-end design that avoids complex preprocessing, postprocessing, and multiple inference stages, its speed significantly outperforms non-end-to-end algorithms. In comparison with existing methods, our MolParser achieves a significantly better speed-accuracy Pareto curve.

In our study, we also conducted a qualitative evaluation of MolParser and found it to be highly robust against noise present in various real-world data. It demonstrated strong parsing capabilities for Markush structures and performed well on many complex molecules—cases that have been challenging for previous methods to address. For more details, please refer to the supplementary materials.

\begin{table*}[h]
\centering%
\resizebox{\linewidth}{!}{
\begin{tabular}{lcccccccc}
\toprule
\textbf{Method} & Vision Backbone & Resolution & Param Count & Throughout $\uparrow$ & \begin{tabular}[c]{@{}c@{}}WildMol-10K $\uparrow$\\ (10000)\end{tabular} & \begin{tabular}[c]{@{}c@{}}WildMol-10K-M $\uparrow$\\ (10000)\end{tabular}\\
\hline
\textit{Open-sourced implements} \\
Img2Mol \cite{clevert2021img2mol} & 8-Layer-CNN  & 224*224 & 201M & 0.38 & 24.4 & - \\
MolGrapher \cite{morin2023molgrapher} & Res18 + Res50  & 1024*1024 & 40M & 2.2 & 45.5 & - \\
DECIMER2.7 \cite{rajan2023decimer} & EfcientNet-B3  & 299*299 & 12M & 0.14 & 56.0 & - \\
MolScribe \cite{qian2023molscribe} & Swin-Base  & 384*384 & 88M & 16.5 & 66.4 & - \\
\hline
\textit{Our end-to-end Molparser} \\
MolParser-Tiny & Swin-Tiny & 224*224 & 66M & \textbf{131.6} & 73.1 & 15.3\\
MolParser-Small & Swin-Small & 224*224 & 108M & 116.3 & 76.3 & 34.8\\
\textbf{MolParser-Base} & Swin-Base & 384*384 & 216M & 39.8 & \textbf{76.9} & \textbf{38.1}\\
MolParser-InternVL & InternViT-300M \cite{miniinternvl} & 448*448 & 2200M & 1.5 & 72.9 & 33.7 \\
\bottomrule
\end{tabular}}\vspace{-1mm}
\caption{\textbf{Speed and accuracy evaluation in WildMol.} We report the throughout and accuracy in our WildMol benchmark. Throughput is measured on a single RTX 4090D GPU, and the time for preprocessing and postprocessing is also included in the calculation. Except for our MolParser model, existing models do not support the evaluation of extreme complex Markush data in WildMol-M Benchmark.}
\label{tab:size}
\end{table*}

\subsection{Ablation study}
\label{ablation}

\textbf{The importance of large scale training data.} Before our MolParser-7M, the largest paired molecule recognition open-source dataset was MolGrapher-300k \cite{morin2023molgrapher}, which included 300k artificially generated molecular images. We used the same model architecture and training methods of MolParser.  As shown in Table \ref{tab:ablation_data}, When we switched our pre-trained dataset to the significantly smaller MolGrapher, there was a noticeable drop in performance. It demonstrates the essential importance of scaling training data for end-to-end OCSR models.

\textbf{The importance of fine-tuning in real data.} As shown in Table \ref{tab:ablation_data}. We demonstrate the effectiveness of our data engine in this study. We compared the performance of MolParser before and after fine-tuning with data obtained from the data engine. We find that training our end-to-end MolParser models solely on synthetic dataset do not yield satisfactory results across various benchmarks. The reason is that the data distribution of the benchmark differs significantly in style from the molecular images generated by RDKit. However, after incorporating real data from our data engine, a significant performance improvement is achieved. demonstrating that the real-world and in-the-wild data extracted through our active learning algorithm is essential.

\begin{table}[h]
\centering%
\resizebox{\linewidth}{!}{
\begin{tabular}{lcccccc}
\toprule
Training Dataset & Fine-tuning & WildMol-10K $\uparrow$ \\ \hline
MolGrapher-300k & - & 22.4  \\
MolParser-7M (pt) & - & 51.9  \\
MolParser-7M (pt+ft) & - & 75.9  \\
MolParser-7M (pt) & MolParser-7M (ft) & \textbf{76.9}  \\
\bottomrule
\end{tabular}}\vspace{-1mm}
\caption{\textbf{Ablation study in training and finetuning dataset.} We report the accuracy score in WildMol-10K benchmarks. 'pt' means the synthetic pretraining subset and 'ft' stand for fine-tuning subset, suggested in Section \ref{sec:datasets}
\label{tab:ablation_data}.}
\end{table}

\textbf{The impact of model scale.} In our study, we experiment with varying input resolutions of image sizes, the quantity of parameters in visual backbones, and the parameter count in transformer decoders. As shown in Table \ref{tab:size}, it demonstrates that scaling model dimensions has a certain effect, yet it is less effective compared to scaling the dataset and fine-tuning in real data from our data engine. Additionally, we observe that employing excessively large end-to-end image caption models, such as Mini-InternVL \cite{miniinternvl}, may render the training process more challenging.

\textbf{The impact of data augmentation.} We run controlled experiments. Table \ref{tab:ablation_aug} reports the results. Data augmentation improves generalization on real scanned benchmarks. Curriculum learning helps further. It starts with weak augmentation and gradually increases strength during training.

\begin{table}[h]
\centering%
\resizebox{0.99\linewidth}{!}{
\begin{tabular}{lcccccc}
\toprule
Data Augmentation & Curriculum Strategy &  WildMol-10K $\uparrow$ \\ \hline
$\times$ & $\times$ & 40.1  \\
$\checkmark$ & $\times$ & 69.5 \\
$\checkmark$ & $\checkmark$ & \textbf{76.9}  \\
\bottomrule
\end{tabular}}\vspace{-1mm}
\caption{\textbf{Ablation study in data augmentation and training strategy.} We report the accuracy score in WildMol-10K benchmarks.}
\label{tab:ablation_aug}
\end{table}

\subsection{Expanding applications: molecular property prediction}
\label{downstream}

\begin{table*}[!ht]
\centering%
\resizebox{0.78\linewidth}{!}{
\begin{tabular}{lccccccc}
\toprule
\textbf{Method} & BBBP $\uparrow$ & Tox21 $\uparrow$ & ToxCast $\uparrow$ & BACE $\uparrow$ & SIDER $\uparrow$ & \textbf{Avg.} $\uparrow$ \\ \hline
\textit{3D Conformation} \\
GEM\cite{gem} & 72.4 & 78.1 & - & \underline{85.6} & \underline{67.2} & - \\
3D InfoMax\cite{3dinfomax} & 68.3 & 76.1 & 64.8 & 79.7 & 60.6 & 69.9 \\
GraphMVP\cite{graphmvp} & 69.4 & 76.2 & 64.5 & 79.8 & 60.5 & 70.1 \\
MoleculeSDE\cite{moleculesde} & 71.8 & 76.8 & 65.0 & 79.5 & \textbf{75.1} & 73.6 \\
Uni-Mol\cite{unimol} & 71.5 & 78.9 & 69.1 & 83.2 & 57.7 & 72.1 \\
MoleBlend\cite{molbend} & \underline{73.0} & 77.8 & 66.1 & 83.7 & 64.9 & 73.1\\
Mol-AE\cite{molae} & 72.0 & \textbf{80.0} & \underline{69.6} & 84.1 & 67.0 & \textbf{74.5}\\
UniCorn\cite{unicorn} & \textbf{74.2} & \underline{79.3} & 69.4 & \textbf{85.8} & 64.0 & \textbf{74.5}\\ \hline
\textit{2D Graph} \\
AttrMask\cite{attrmask} & 65.0 & 74.8 & 62.9 & 79.7 & 61.2 & 68.7\\
GROVER\cite{grover} & 70.0 & 74.3 & 65.4 & 82.6 & 64.8 & 71.4\\
BGRL\cite{bgrl} & 72.7 & 75.8 & 65.1 & 74.7 & 60.4 & 69.7\\
MolCLR\cite{molclr} & 66.6 & 73.0 & 62.9 & 71.5 & 57.5 & 66.3\\
GraphMAE\cite{graphmae} & 72.0 & 75.5 & 64.1 & 83.1 & 60.3 & 71.0\\
Mole-BERT\cite{molbert} & 71.9 & 76.8 & 64.3 & 80.8 & 62.8 & 71.3\\
SimSGT\cite{simsgt} & 72.2 & 76.8 & 65.9 & 84.3 & 61.7 & 72.2\\
MolCA + 2D\cite{molca} & 70.0 & 77.2 & 64.5 & 79.8 & 63.0 & - \\ \hline
\textit{2D Image} \\
Swin-T (w/ ImageNet pretrained) & 62.5 & 77.9 & 67.4 & 76.0 & 60.5 & 68.9\\
\textbf{Swin-T (w/ MolParser pretrained)} & 70.4 & 79.0 & \textbf{74.6} & 84.1 & 60.2 & \underline{73.7}\\
\bottomrule
\end{tabular}}\vspace{-1mm}
\caption{\textbf{Comparison of molecular property prediction methods.} We report the average ROC-AUC scores after five runs.}
\label{tab:molnet}
\end{table*}

We make the unexpected observation that the image feature extractor of our MolParser, a Swin Transformer, can serve as an effective molecular fingerprint (or molecular embedding) for downstream molecular property prediction tasks after being trained on the MolParser-7M dataset. Specifically, for each molecule, we first use the RDKit toolkit \cite{rdkit} to render a 2D molecule structural image, then extract visual features using the MolParser vision backbone, and apply global average pooling to obtain a 2048-dimensional feature vector as the molecular representation. A simple two-layer MLP is then used to perform molecular property prediction.

We evaluate our approach alongside several baselines on five molecular property classification tasks from the MoleculeNet benchmark \cite{wu2018moleculenet}. As shown in Table \ref{tab:molnet}, our method achieves competitive performance in molecular property prediction. Notably, features extracted by MolParser, when paired with a lightweight MLP, perform on par with more complex models that rely on 2D or 3D graph-based representations. Moreover, our approach substantially outperforms other image-based feature extractors that are not pre-trained on the MolParser-7M dataset. These results indicate that 2D molecular structure images contain rich, chemically meaningful information. They also highlight the effectiveness of our end-to-end, large-scale OCSR training in learning high-quality visual representations for downstream chemical tasks.

\subsection{Expanding applications: chemical reaction parsing}
\label{downstream2}

Following OmniParser \cite{lu2024omniparser}, we input the molecular locations detected by our MolDet model and the E-SMILES sequences recognized by our MolParser model into GPT-4o \cite{gpt4o} to enhance chemical reaction parsing. We draw bounding boxes around target molecules in chemical reaction images and label each with a corresponding molecule index, which is provided as input to GPT-4o. Additionally, we prepend the prompt with the E-SMILES recognized for each indexed molecule. This significantly enhances the ability of MLLMs like GPT-4o to process chemical reaction images. Refer to the appendix for details.
\section{Conclusion}
\label{sec:conclusion}

We propose a novel end-to-end OCSR algorithm that extends the SMILES representation and introduces a large-scale training dataset, MolParser-7M. The model is pretrained on large scale synthesized data and fine-tuned on manually annotated in-the-wild samples. It outperforms existing methods on classical OCSR benchmarks and our newly introduced WildMol benchmark. The system achieves high processing speed and strong performance in extracting structured molecular information from real, unstructured scientific literature.

Despite its effectiveness, MolParser still has room for improvement. For instance, molecular chirality, which is closely related to chemical properties, is not yet fully exploited. In addition, scaling up the amount of real annotated training data may further boost performance. As future work, we aim to address the challenge of chirality in OCSR and scale up the volume of real-world training data. We also plan to use MolParser to extract molecules and Markush structures with their visual fingerprints from large-scale scientific literature and patents, enabling the creation of a comprehensive database for chemical information mining.
{
    \small
    \bibliographystyle{ieeenat_fullname}
    \bibliography{main}
}

\clearpage
\setcounter{page}{1}
\maketitlesupplementary

\section{Open Source Materials}
\label{sup:opensource}

MolParser-7M dataset is open sourced in \href{https://huggingface.co/datasets/AI4Industry/MolParser-7M}{HuggingFace Dataset}. The yolo11 model used for detection molecule structure is also avaliable in \href{https://huggingface.co/AI4Industry/MolDet}{HuggingFace Model}. We also provide a \href{https://ocsr.dp.tech/}{OCSR demo} using our MolParser-Base model.

\section{Extended SMILES Explanation}
\label{sup:extended_smiles}

The extended SMILES format is defined as:

\begin{center}
    \texttt{SMILES<sep>EXTENSION}
\end{center}

\begin{enumerate}
    \item \texttt{SMILES} represents an RDKit-compatible SMILES expression. Each molecule has a unique representation that can be generated (for non-Markush molecules) using the following method, where \texttt{rootedAtAtom=0} indicates that the SMILES generation starts from the atom indexed at 0.
    \item \texttt{<sep>} is the delimiter separating the RDKit-compatible SMILES string from its extended description. The part before the delimiter is the RDKit-compatible SMILES, while the part after provides supplemental information (e.g., Markush groups, connection points, repeating groups).
    \item \texttt{EXTENSION} is an optional component that supplements the preceding SMILES with descriptions written in XML format, including groups surrounded by special tokens of three types:
    \begin{enumerate}
        \item \texttt{<a>[ATOM\_INDEX]:[GROUP\_NAME]</a>} indicates a substituent.
        \item \texttt{<r>[RING\_INDEX]:[GROUP\_NAME]</r>} represents a group connected at any position of a ring.
        \item \texttt{<c>[CIRCLE\_INDEX]:[CIRCLE\_NAME]</c>} denotes abstract ring.
    \end{enumerate}
    An additional special token \texttt{<dum>} indicates a connection point.

    \textbf{Definitions:}
    \begin{itemize}
        \item \texttt{ATOM\_INDEX} refers to the atom index at which the substituent is located (starting from 0).
        \item \texttt{RING\_INDEX} denotes the ring index (starting from 0).
        \item \texttt{GROUP\_NAME} specifies the name of the substituent, which can be an abbreviated group, general substituent, or Markush group, such as \texttt{R, X, Y, Z, Ph, Me, OMe, CF3}, etc. It may also be \texttt{<dum>} to indicate a connection point. For Markush substituents with superscripts or subscripts, these can be represented within square brackets, e.g., \texttt{R[1], R[3]}.
        \item \texttt{CIRCLE\_INDEX} refers to the index of the named ring (starting from 0).
        \item \texttt{CIRCLE\_NAME} indicates the name of the ring.
    \end{itemize}
\end{enumerate}

Figure \ref{fig:smiles_example} shows the usage of extended SMILES:

\begin{figure}[h]
    \centering
    \includegraphics[width=0.1\textwidth]{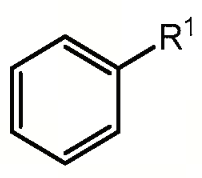}
    \caption*{\texttt{*c1ccccc1<sep>\textcolor{red}{<a>0:R[1]</a>}}}

    \vspace{5pt} 
    \includegraphics[width=0.13\textwidth]{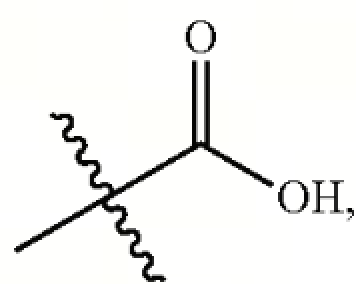}
    \caption*{\texttt{*C(O)=O<sep>\textcolor{red}{<a>0:<dum></a>}}}

    \vspace{5pt}
    \includegraphics[width=0.2\textwidth]{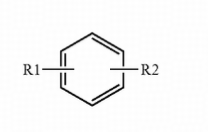}
    \caption*{\texttt{c1ccccc1<sep>\textcolor{red}{<r>0:R[1]</r>}<r>0:R[2]</r>}}

    \vspace{5pt}
    \includegraphics[width=0.25\textwidth]{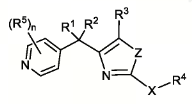}
    \caption*{\texttt{**C1*C(*)=C(C(*)(*)C2=CC=NC=C2)N=1<sep>\\<a>0:R[4]</a><a>1:X</a>\textcolor{red}{<r>1:R[5]?n</r>}\\<a>3:Z</a><a>5:R[3]</a><a>8:R[2]</a><a>9:R[1]</a>}}

    \vspace{5pt}
    \includegraphics[width=0.3\textwidth]{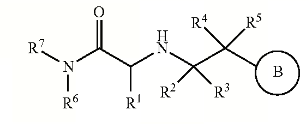}
    \caption*{\texttt{*C(NC(*)(*)C(*)(*)*)C(=O)N(*)*<sep>\\<a>0:R[1]</a><a>4:R[3]</a><a>5:R[2]</a><a>7:R[5]</a>\\<a>8:R[4]</a>\textcolor{red}{<c>0:B</c>}<a>13:R[7]</a><a>14:R[6]</a>}}
    \captionsetup{justification=raggedright, singlelinecheck=false}
    \caption{Molecule images examples with extended SMILES. The red parts are as follows: Markush group, attachment point, ring attachment with uncertainty position, duplicated structure and abstract ring.}
    \label{fig:smiles_example}
\end{figure}

\begin{figure*}[h]
    \centering  
    \begin{minipage}[t]{\textwidth}  
        \centering
        \includegraphics[width=\textwidth]{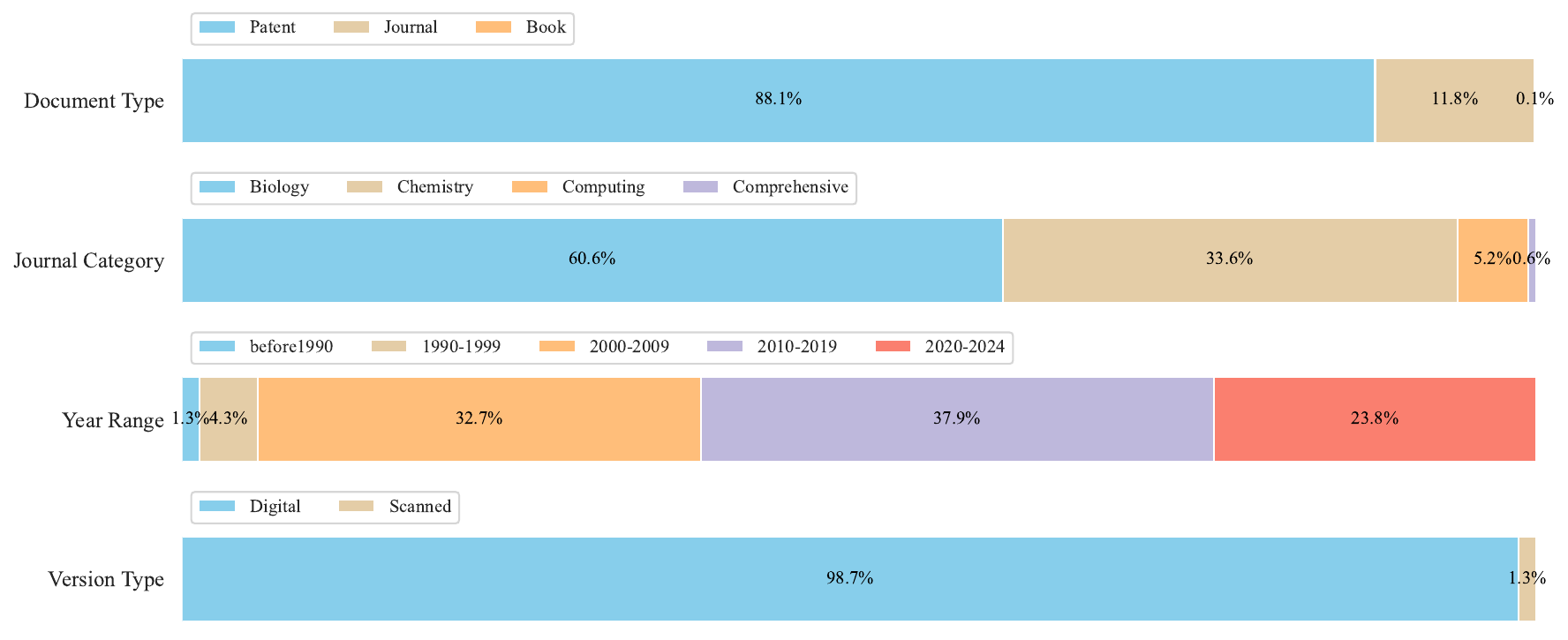}
    \end{minipage}
    \vfill
    \begin{minipage}[t]{0.9\textwidth} 
        \hspace{0.02\textwidth}  
        \includegraphics[width=\textwidth]{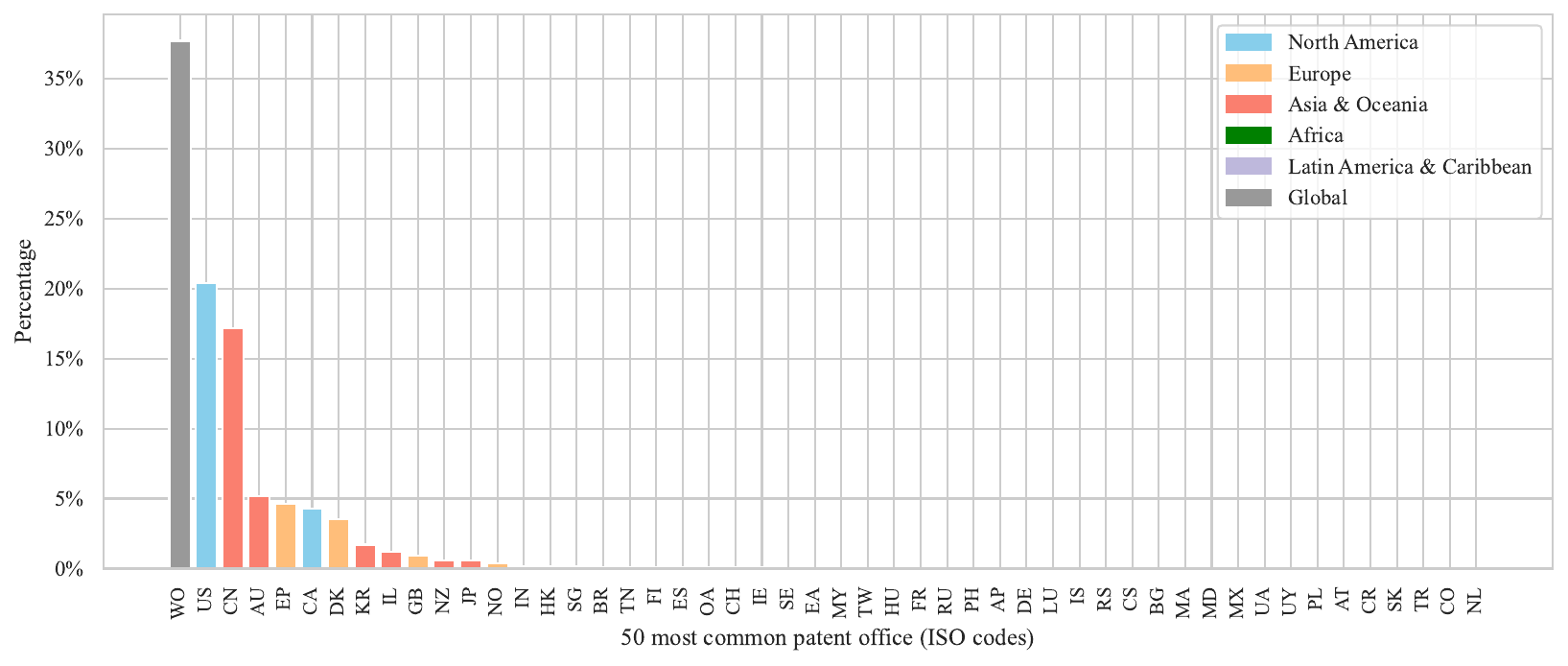}
    \end{minipage}
    \caption{\textbf{Statistical analysis of MolParser-SFT original PDF database.} We compiled source information for the collected PDF files, including article type, publication date, PDF format, journal subject distribution, and patent office sources.}
    \label{fig:dataset_pdf}
\end{figure*}

\begin{figure*}[h]
    \begin{minipage}[b]{0.49\textwidth}
        \centering
        \includegraphics[width=\textwidth]{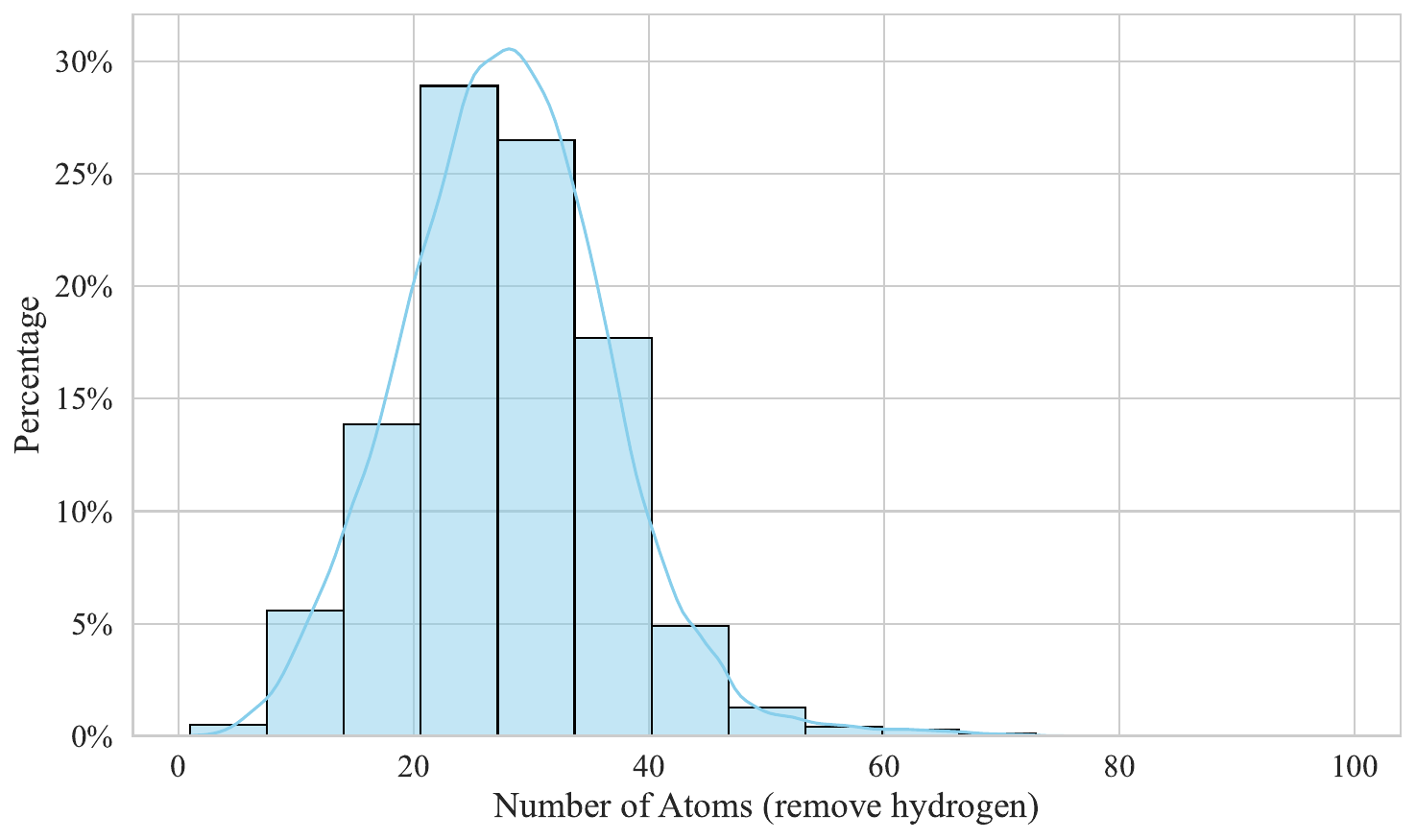}
        \subcaption{Distribution of atom count} \label{fig:sub1}
    \end{minipage}%
    \hfill
    \begin{minipage}[b]{0.49\textwidth}
        \centering
        \includegraphics[width=\textwidth]{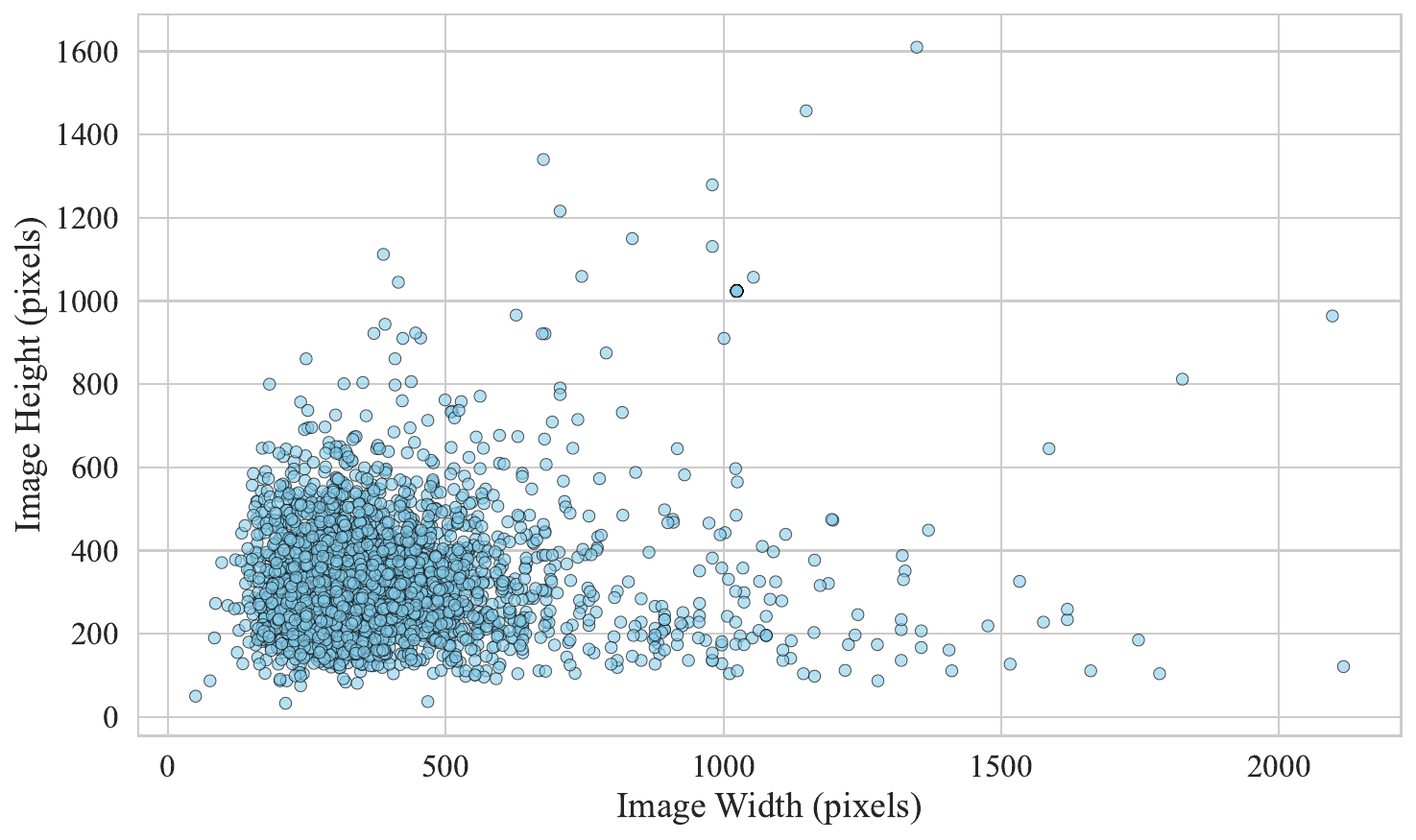}
        \subcaption{Distribution of image size} \label{fig:sub2}
    \end{minipage}
    \caption{\textbf{Statistical analysis of MolParser-7M dataset.} The distribution of molecule atom counts and molecule image sizes. Compared to the fixed-size synthetic datasets used in other studies, our dataset exhibits a wider range of image sizes and aspect ratios.}
    \label{fig:dataset_all}
\end{figure*}

\section{Dataset}
\label{sup:dataset}

\subsection{Statistical information of MolParser-7M}

Molparser-7M contains a total of 7,740,871 paired OCSR training data, making it the largest open-source paired OCSR dataset currently available. It is important to note that the open-source datasets MolGrapher-300k and Img2Mol are both subsets of our Molparser-7M. Additionally, as shown in Figure \ref{fig:dataset_pdf} and \ref{fig:dataset_all}, the data distribution of our MolParser is more comprehensive.

\subsection{Data Augmentation}
\textbf{Render Augmentation} 
During synthetic data generation in our data engine, we incorporat several augmentations for rendering molecular structure diagrams, similar to those used in MolGrapher \cite{morin2023molgrapher}. Augmentations such as bond width, font type, font size, rotation, and aromatic cycle representation are randomly applied during rendering.

\textbf{Image Augmentation}
Whether for synthetic data or real data, we also apply image augmentations during training. We use several types of data augmentation, including RandomAffine, JPEGCompress, InverseColor, SurroundingCharacters, RandomCircle, ColorJitter, Downscale and Bounds. These type of augmentation are visualized in Figure \ref{fig:augment}.

\textbf{SMILES Augmentation}
We apply SMILES augmentation only during pre-training. Since a molecule's SMILES representation varies with the choice of root atom, we randomly change the root atom to help the transformer learn SMILES syntax more robustly. During fine-tuning, augmentation is disabled and the root atom is fixed to index zero, reducing ambiguity during generation.

\section{Experiment Setting}
All variants of MolParser adopt a BART decoder with 12 transformer decoder layers and 16 attention heads. An MLP connector reduces the channel dimension of the visual encoder output by half. The Swin Transformer produces a feature map of size bs x n × n × d, which is flattened into a sequence and used as prefix tokens for the decoder.

During the pre-training stage, we train the model for 20 epochs using the AdamW optimizer with a learning rate of 1e-4, a weight decay of 1e-2, and a cosine learning rate schedule with warmup. We set label smoothing to 0.01. In the fine-tuning phase, we reduce the learning rate to 5e-5, decrease the number of training epochs to 4, and lower the label smoothing to 0.005. All experiments are conducted on 8 NVIDIA RTX 4090D GPUs.

\section{Case Study}
\begin{figure*}[h]
    \centering
    \includegraphics[width=0.65\textwidth]{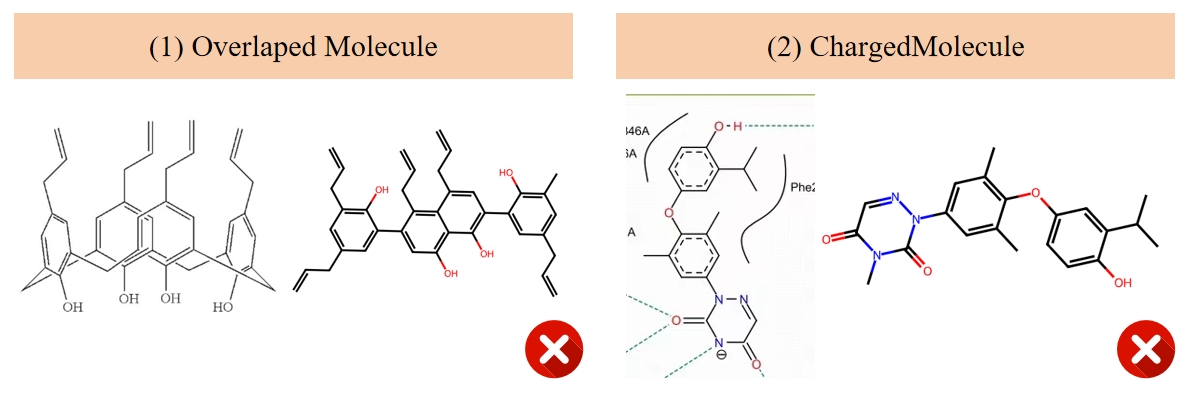}
    \caption{\textbf{MolParser failure case.} The figure shows the broad diversity of predictions made by MolParser for input molecular images. The input image (left) is displayed alongside the predicted molecule rendered by E-SMILES prediction (right).}
    \label{fig:badcase}
\end{figure*}

\begin{figure*}[!t]
    \centering
    \includegraphics[width=\textwidth]{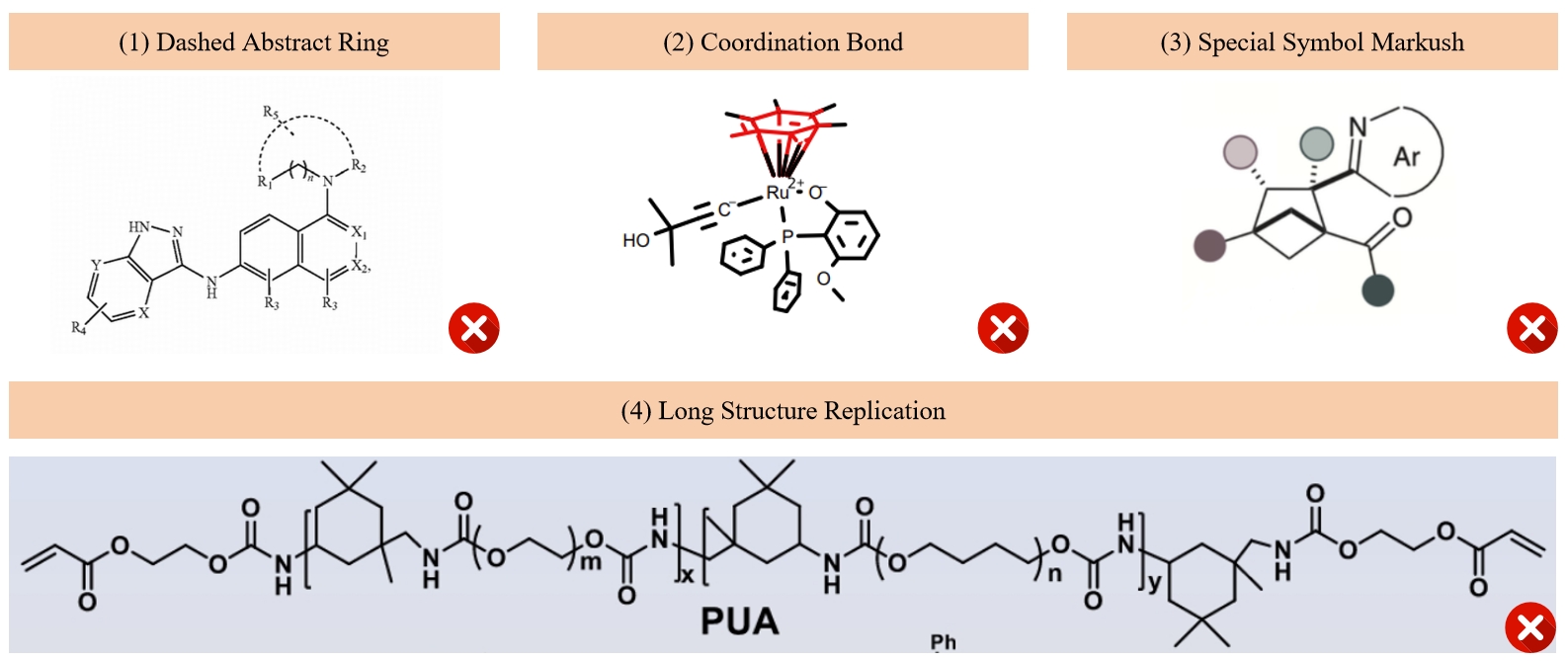}
    \caption{\textbf{E-SMILES failure case.} Molecular structures with dashed lines representing abstract rings, structures with coordination bonds, and Markush structures depicted using special patterns are not currently supported in E-SMILES notation. Additionally, the replication of long structural segments on the skeleton, rather than individual atoms, is also not supported by our E-SMILES format.}
    \label{fig:smibadcase}
\end{figure*}

We test numerous examples and conduct qualitative analysis. Our MolParser performs well on stylized and low-quality molecular images that challenge previous algorithms, though it struggles with overlapped large molecules, charged molecules, and some cases where E-SMILES cannot effectively represent the structures, which shown in figure \ref{fig:badcase} and figure \ref{fig:smibadcase}.

\section{Downstream Usage}
\label{sup:downstream}

In unstructured documents, extracting molecular structures and leveraging LLMs for structured information extraction has become a key application of Optical Chemical Structure Recognition (OCSR). We first convert each PDF page into an image and use a YOLO11 \cite{yolo11} model to detect molecular structures. The detected molecules are then parsed by our MolParser and converted into an extended, XML-like SMILES format that is more LLM-friendly. This representation allows LLMs to easily identify which groups undergo transformations in chemical reactions. Following OmniParser \cite{lu2024omniparser}, we integrate molecular location and SMILES information into GPT-4o \cite{gpt4o} to enhance MolParser’s ability to parse full chemical reaction formulas.

\begin{figure}[!h]\centering
\begin{minipage}{0.95\linewidth}\vspace{0mm}    
    \centering
    \scriptsize
    \begin{tcolorbox}[boxrule=0.2mm]
        \centering
        \hspace{-5mm}
        \begin{tabular}{p{0.99\columnwidth}}
        \hspace{1mm}
        \begin{minipage}{0.99\columnwidth}
        
        \includegraphics[width=\linewidth]{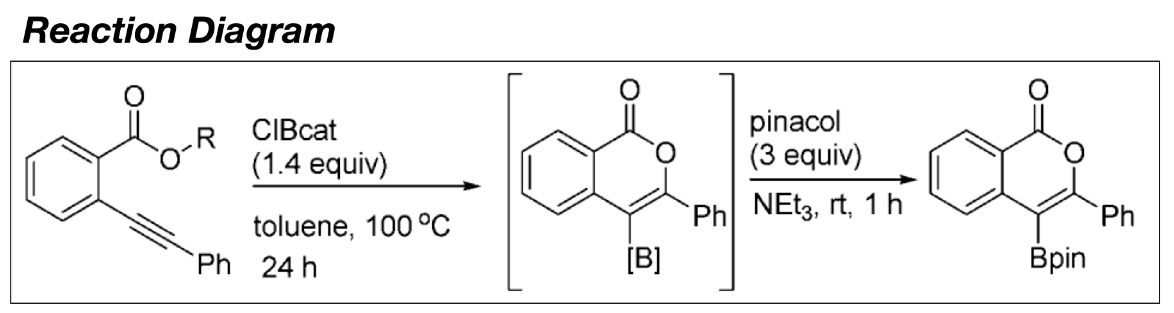}

        \textbf{User:} Explain this reaction in detail.
        \vspace{4px}

        \textbf{MolParser:} Applying molecule detection and OCSR.
        \vspace{4px}

        \includegraphics[width=\linewidth]{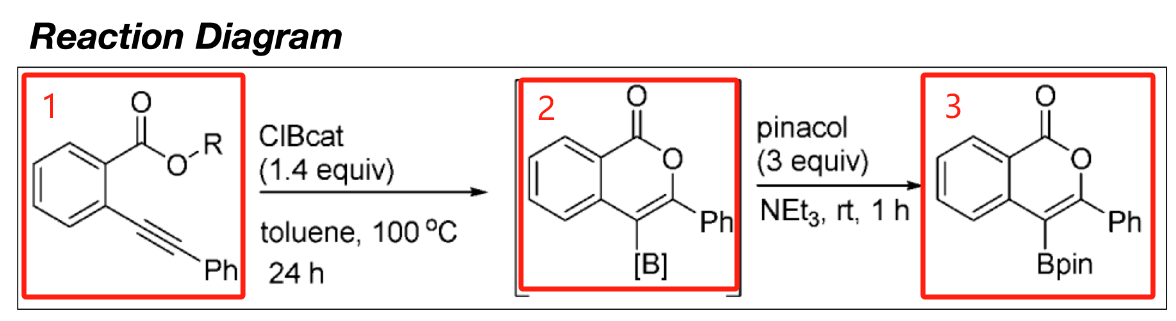}

        \textbf{MolParser:} \\
        Extended SMILES of Molecules in Image:\\
        Molecule Box ID 1: \texttt{*OC(=O)C1=CC=CC=C1C\#C*<sep>\\<a>0:R</a><a>12:Ph</a>}\\
        Molecule Box ID 2: \texttt{O=c1c2c(cccc2)c(*)c(*)o1<sep>\\<a>9:B</a><a>11:Ph</a>}\\
        Molecule Box ID 3: \texttt{O=c1oc(*)c(*)c2ccccc12<sep>\\<a>4:Ph</a><a>6:Bpin</a>}
        \vspace{4px}

        \textbf{GPT-4o:} \\
        This reaction diagram illustrates a two-step synthetic pathway:\\
        First Step: Starting from molecule \\
        \texttt{*OC(=O)C1=CC=CC=C1C\#C*<sep>\\<a>0:R</a><a>12:Ph</a>}, \\
        which has an alkyne and ester group, the reaction proceeds with ClBcat (1.4 equivalents) in toluene at 100 °C for 24 hours. This yields the intermediate molecule \\
        \texttt{"O=c1c2c(cccc2)c(*)c(*)o1<sep>\\<a>9:B</a><a>11:Ph</a>"}, \\
        which has a fused aromatic ring with a carbonyl group.\\
        Second Step: The intermediate \\
        \texttt{O=c1c2c(cccc2)c(*)c(*)o1<sep>\\<a>9:B</a><a>11:Ph</a>}\\
        is then reacted with pinacol (3 equivalents) and $NEt_3$ at room temperature for 1 hour, producing the final product \texttt{O=c1oc(*)c(*)c2ccccc12<sep>\\<a>4:Ph</a><a>6:Bpin</a>},\\
        where a Bpin group is introduced onto the fused ring structure.
        \vspace{4px}
        
        \end{minipage}
        \end{tabular}
    \end{tcolorbox}
\end{minipage}
\label{fig:qa}
\end{figure}

\begin{figure*}[h]
    \centering
    \includegraphics[width=\textwidth]{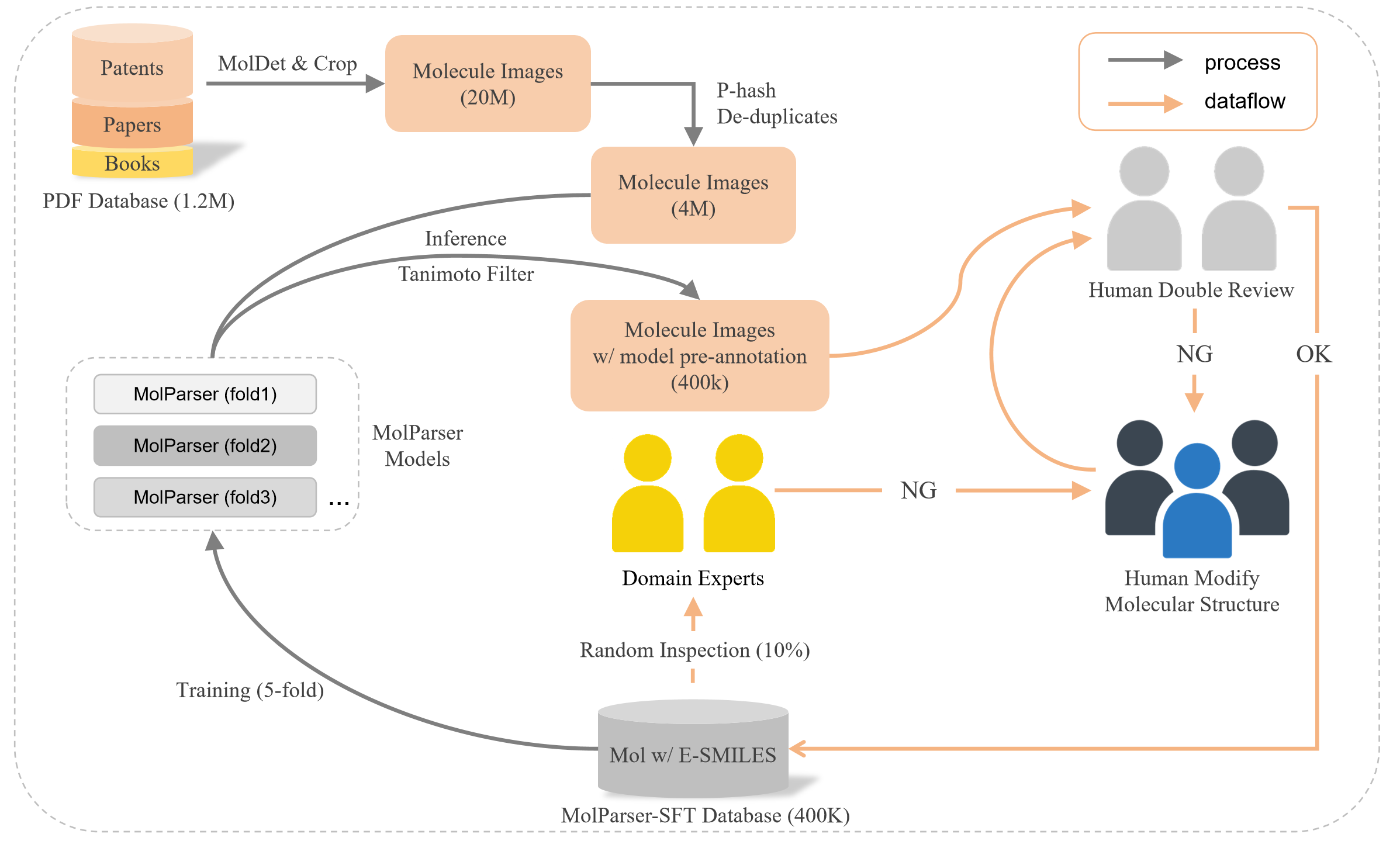}
    \caption{\textbf{MolParser data engine.} We design a human-in-the-loop active learning framework, using Tanimoto similarity scores of multiple model predictions to select molecules for training. Each molecule image is pre-labeled by the model, reviewed by two annotators, and subject to expert inspection.}
    \label{fig:augment}
\end{figure*}

\begin{figure*}[h]
    \centering
    \includegraphics[width=0.65\textwidth]{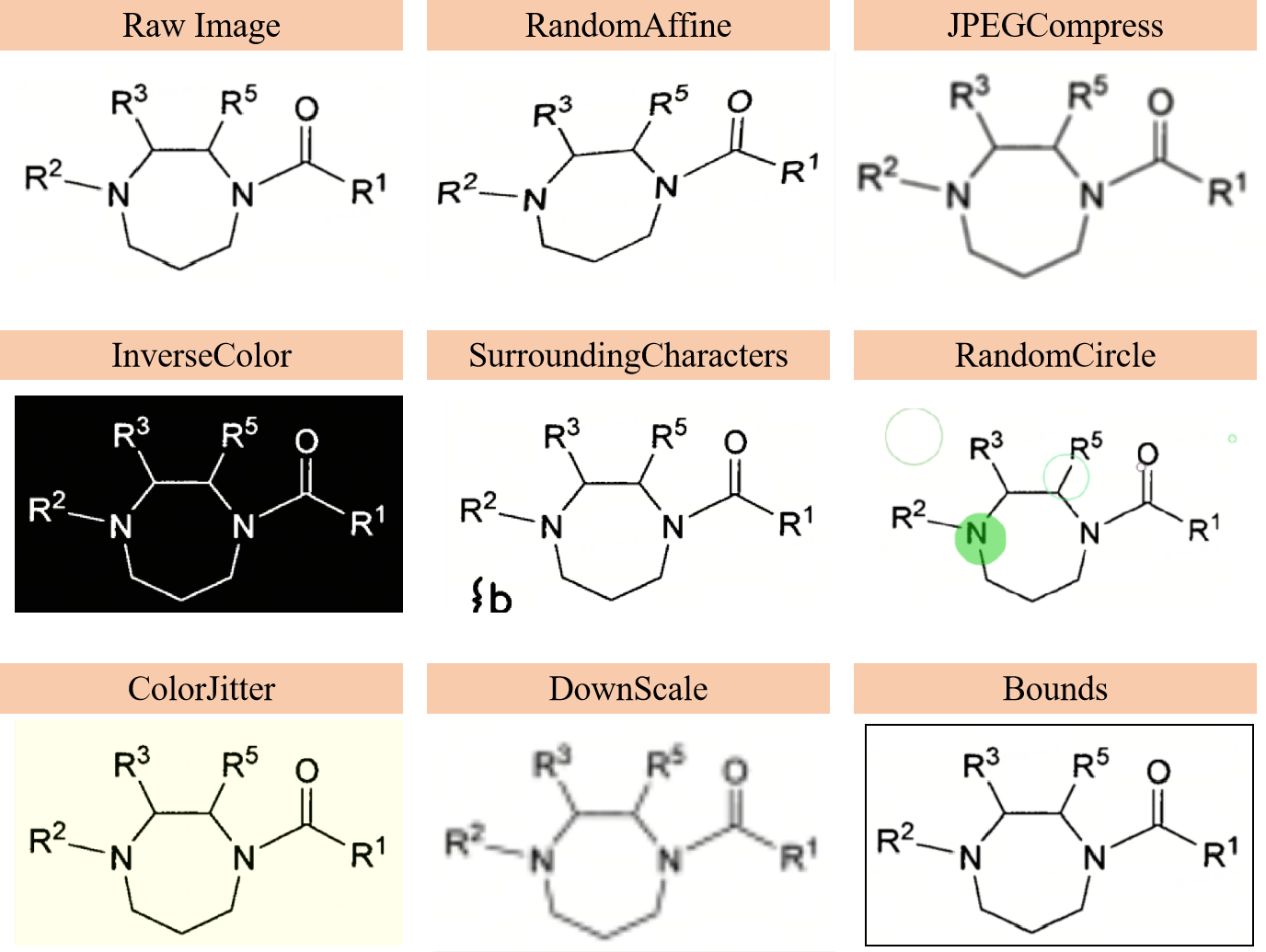}
    \caption{\textbf{Data augmentation in training.} We design the augm entation of the image according to the noise that may occur in real data, which cropped from scanned PDF files.}
    \label{fig:augment}
\end{figure*}

\begin{figure*}[h]
    \centering
    \includegraphics[width=0.65\textwidth]{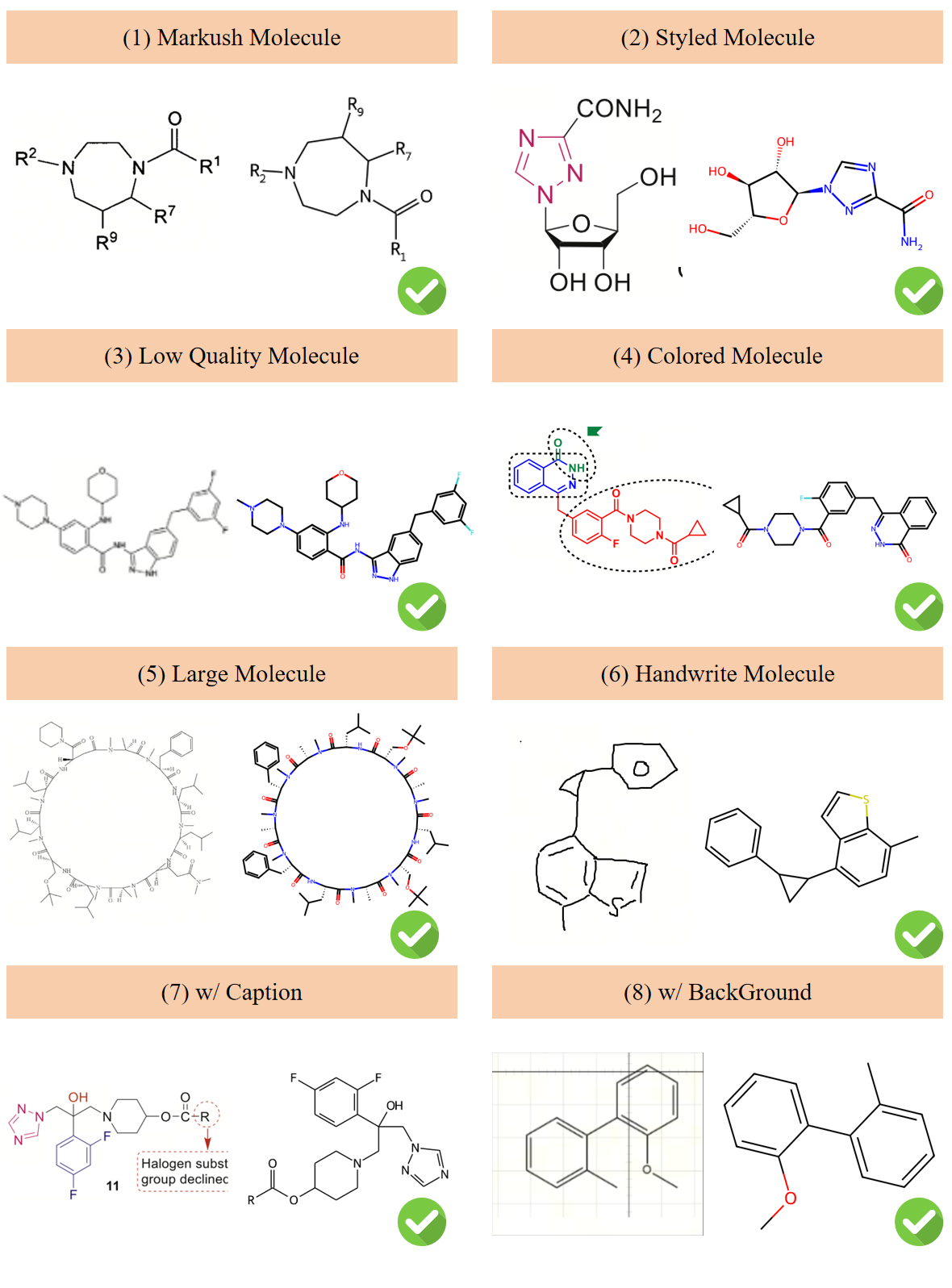}
    \caption{\textbf{MolParser qualitative evaluation.} The figure shows the broad diversity of predictions made by MolParser for input molecular images. The input image (left) is displayed alongside the predicted molecule rendered by E-SMILES prediction (right).}
    \label{fig:example}
\end{figure*}

\begin{figure*}[h]
    \centering
    \includegraphics[width=0.9\textwidth]{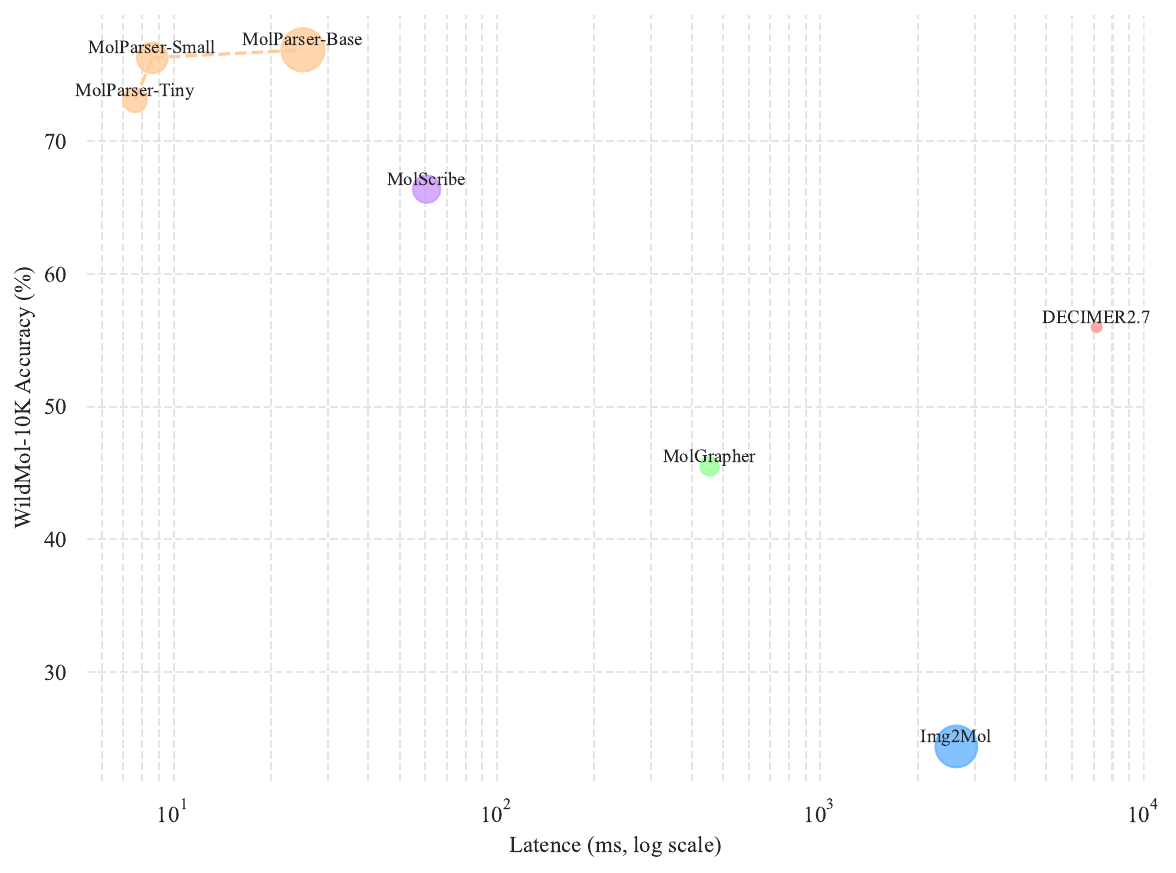}
    \caption{\textbf{The speed-accuracy Pareto curve of the OCSR system.} Models toward the top-left corner are better. The size of the circles represents the model's parameter count, and the time is tested on a single RTX-4090D GPU for the entire pipeline.}
    \label{fig:pareto}
\end{figure*}

\end{document}